\documentclass{article}

\usepackage[utf8]{inputenc} 
\usepackage[T1]{fontenc}    
\usepackage{hyperref}       
\usepackage{url}            
\usepackage{booktabs}       
\usepackage{amsfonts}       
\usepackage{nicefrac}       
\usepackage{microtype}      

\usepackage{amsmath}
\usepackage{bm}
\usepackage{graphicx}
\usepackage{subcaption}

\usepackage{multirow}

\usepackage{geometry}
\usepackage{bm}    
\usepackage{amssymb}
\usepackage{authblk}
\usepackage{float}
\usepackage{afterpage}

\title{Recent Progress of Face Image Synthesis}

\author{Zhihe Lu, Zhihang Li, Jie Cao, Ran He and Zhenan Sun\\
National Laboratory of Pattern Recognition, CASIA \\
Center for Research on Intelligent Perception and Computing, CASIA \\
Center for Excellence in Brain Science and Intelligence Technology, CAS\\
University of Chinese Academy of Sciences, Beijing, 100049, China\\
{\tt\small \{luzhihe2016, lizhihang2016, caojie2016\}@ia.ac.cn, \{rhe, znsun\}@nlpr.ia.ac.cn}
}

\begin{document}
\date{}
\maketitle

\begin{abstract}
\noindent Face synthesis has been a fascinating yet challenging problem in computer vision and machine learning. Its main research effort is to design algorithms to generate photo-realistic face images via given semantic domain. It has been a crucial prepossessing step of main-stream face recognition approaches and an excellent test of AI ability to use complicated probability distributions. In this paper, we provide a comprehensive review of typical face synthesis works that involve traditional methods as well as advanced deep learning approaches. Particularly, Generative Adversarial Net (GAN) is highlighted to generate photo-realistic and identity preserving results. Furthermore, the public available databases and evaluation metrics are introduced in details. We end the review with discussing unsolved difficulties and promising directions for future research.
\end{abstract}

\section{Introduction}

As one of the most successful applications of image analysis and understanding, face synthesis has recently received significant attention, especially during the last decade. The major purpose of face synthesis is to synthesize desired images, e.g.\ photo-realistic, some artistic style, or clearer faces via given input in some semantic domains. In this paper, traditional methods as well as advanced deep learning approaches for face synthesis will be extensively reviewed and discussed.

Non-neural-network approaches are summarized as traditional methods which are divided into three broad categories, i.e.\ subspace representation, geometry modeling and statistical models. Early researchers mainly focused on struggle with the curse of dimensionality, e.g.\ Huang et al.\ \cite{huang2010manifold} utilized manifold estimation, and \cite{qiao2010tensor, zhang2014face} proposed tensor-based subspace learning method. Some people deployed the mixture of several models \cite{heo2012gender, heo20123} to handle the problem. Inspired by massive successful works in computer vision from geometry knowledge, Halder et al.\ \cite{halder2010face} provided a face synthesis system based on geometry modeling and Patel and Smith \cite{patel2012driving} used the information of the obscure part as auxiliary condition to obtain 3D morphable model. In order to build or expand the database for face recognition, a method by Ramalingam and Viet \cite{ramalingam20133d} focused on using inexpensive 3D sensors to synthesize face and Song and Zhang \cite{yuhao2016framework} analyzed the tensor's mathematical properties of face to generate the novel face images. Simultaneously, a bunch of works have been proposed with statistical learning theory.

Deep learning has achieved great breakthroughs in image classification \cite{krizhevsky2012imagenet}, image generation \cite{zhmoginov2016inverting,richardson20163d}, biometrics \cite{taigman2014deepface,yi2014learning,peng2017reconstruction,zhu2014multi}, etc. Particularly, convolution neural network dose remarkably well in tasks related to image \cite{taigman2014deepface,yi2014learning,peng2017reconstruction,richardson20163d}. A tremendous amount of works based on deep learning have been developed in face synthesis. Peng et al.\ \cite{peng2017reconstruction} concentrated on a synthesis CNN to generate non-frontal view from a single frontal face and Richardson et al.\ \cite{richardson20163d} extracted the face geometry from its image directly by a CNN based approach. As an extention of traditional AAM models, Duong et al.\ \cite{duong2016deep} proposed a Deep Appearance Models (DAMs) approach, which uses Deep Boltzmann Machines (DBM) to robustly capture the variations of facial shapes and appearances. 

Goodfellow et al.\ \cite{gan} proposed the conception of Generative Adversarial network (GAN) in 2014, and later several successful revamped versions \cite{wgan,ebgan} improved the stability of training process. Researchers fused the information and prior knowledge in face synthesis field and produced fake images that almost have no distinction with real ones if observed by human. Meanwhile, a lot of fascinating adapted GAN models have been brought up to deal with plenty of applications in face synthesis. Till now, some state-of-the-art results produced by deep models are based on the GAN framework.

The remainder of this paper is organized as follows: The traditional methods are discussed in the section \ref{Traditional}. Advanced deep learning approaches and GANs which will be emphasized are introduced in Section \ref{deep}. Some public available face databases and evaluation metrics are shown in Section \ref{database and metrics}. In Section \ref{Conclusion}, we conclude the whole paper with discussing the unsolved problems and further promising research directions.

\section{Traditional methods}\label{Traditional}

Many traditional methods of face synthesis have been proposed, which including subspace representation, geometry modeling and statistical models. In this part, relevant researches as well as their purposes will be discussed.

\subsection{Subspace Representation}


Face image lies in an embedded non-linear manifold within the high-dimensional space \cite{huang2010manifold}, which brings great challenges to face synthesis. Subspace learning based methods tackle this problem by mapping high-dimensional data to a low-dimensional space while retaining as much information as possible \cite{huang2010manifold}. Owing to a basic assumption that data lies in a low-dimensional manifold of a high-dimensional space, face synthesis can be performed by subspace learning. 

While modeling a face directly with a linear subspace model is difficult, layered modeling can simplify the matter. Nguyen et al.\ \cite{nguyen2008image} presented an approach that can extract different layers automatically to support the process of face synthesis. Later, Huang et al.\ \cite{huang2010manifold} proposed a method to synthesize 3D face images from a single image with manifold estimation. One sub-problem, sketch synthesis, is to convert a photo into a sketch. For instance,  Chang et al.\ \cite{chang2010facenew} divided the photos and sketches into patches and computed the sparse representation. In order to decrease the computational complexity, Wang et al.\ \cite{wang2017random} employed a novel method of face sketch synthesis based on random sampling and locality constraint. 

With generating multi-pose images or 3D face images from a single image being a significant research direction, Qiao et al.\ \cite{qiao2010tensor} presented a tensor-based subspace learning method (TSL). Moreover, Gender and Ethnicity Specific GEMs (GE-GEMs) is considered in \cite{heo2012gender}. Heo and Savvides \cite{heo20123} further proposed a 3D Generic Elastic Model (3D-GEM) to model faces utilizing Combination of ASMs and AAMs (CASAAMs) to represent a sparse 2-D shape. Confronted with the problem of data sparsity, Xu and Zha \cite{xu2013manifold} deployed transfer learning approach to generate auxiliary data from original sparse data. It was reported in \cite{zhao2014sparse} that a triangulation-based partition criterion ensures the strict alignment of corresponding triangular patches and sparse representation is utilizd to produce high quality frontal face images. Xu and Savvides \cite{juefei2014facial} proposed the Fukunaga Koontz Transform (FKT) approach to model dual subspace for synthesizing facial ethnic appearances. 

Based on the Lambertian reflection model and linear observation model, Zhang et al.\ \cite{zhang2014face} presented a tensor analysis method to synthesize an artificial high-resolution (HR) visual light (VIS) face image from a low-resolution (LR) near-infrared (NIR) input image. Besides, Dou et al.\ \cite{dou2014robust} formulated the monocular face shape reconstruction problem as a Two-Fold Coupled Structure Learning (2FCSL) process. 

\subsection{Geometry Modeling}

Face image contains rich and complicated structural information resulting from the variations in expressions, poses, textures, illumination, etc, which is the major stumbling block of face synthesis. Considering facial key points, texture, shape and other graphics characteristics from geometric perspective, geometry based methods can model face directly. Moreover, the performance of face synthesis can be significantly improved after obtaining face geometry model.

To parameterize the facial images with multi-modeling is proposed by \cite{zhuang2008expressive} and they aim to synthesize accurate facial expression images. A Face Synthesis system (FASY) is introduced in \cite{halder2010face}. Similarly, Bhattacharjee et al.\ \cite{bhattacharjee2011construction} proposed face construction method to generate expected face image with textual description via stored facial components from distinct databases. Furthermore, Patel and Smith \cite{patel2012driving} utilized shape-from-shading as an auxiliary approach to gain a better fitting 3D morphable model. To create databases for face recognition, Ramalingam and Viet \cite{ramalingam20133d} proposed a method using inexpensive 3D sensors to synthesize face image. Occlusion problem is studied in \cite{zhao2016mask}. In addition, Ferrari et al.\ \cite{ferrari2016effective} extended a 3D based frontalization for synthesizing unconstrained face images. To reconstruct individual 3D shapes from multiple single images of one person, Piotraschke and Blanz \cite{piotraschke2016automated} proposed a quality measure that judges a reconstruction without information about the true shape. A novel data augmentation method is presented in \cite{banerjee2017srefi} that can solve the unbalanced dataset problem. It is reported in \cite{jones2017utility} that multiple synthesized facial images can improve the results of recognizing the unfamiliar faces compared with only a single image.

\subsection{Statistical models}

Statistical learning based models have widely applied in computer vision, machine learning and biometrics, which can discover statistical properties from big data. Lots of effort has been made in face synthesis by statistical models.

Wang et al.\ \cite{wang2009face} proposed a 3D spherical harmonic basis morphable model (SHBMM) that combines spherical harmonics with the morphable model framework. To synthesize face sketch, Wang and Tang \cite{wangxiao2009face} used multi-scale Markov Random Fields to address the shortcomings of the linear processes. The embedded hidden Markov model(E-HMM) \cite{zhong2007facenew}, Bayesian inference \cite{zhang2015facenew}, multi-representation approach \cite{peng2016multiplenew} are proposed to improve effectiveness. An architecture that consists of probabilistic face diffuse model and generic face specular map is presented by \cite{shim2010probabilistic}. Observing frontal facial image is the one having the minimum rank of all different poses, Sagonas et al.\ \cite{sagonas2015robust} employed a novel method for joint frontal view reconstruction. Liu et al.\ \cite{liu20153d} investigated thoroughly such cascaded regression based 3D face reconstruction approaches. Then Liu et al.\ \cite{liu2016joint} further proposed an approach to simultaneously solve face alignment and 3D face reconstruction by iteratively and alternately applying two sets of cascaded regressors. Introducing guided image filters to capture detailed features is proposed in \cite{dang2015detail}. Moreover, Ren et al.\ \cite{ren2016example} achieved face image synthesis by upscaling and refining the estimated image several times. A two-step method for cross-modality face synthesis is proposed in \cite{song2016structured}. Recently, Bordes et al.\ \cite{bordes2017learning} presented a method for training a generative model via an iterative denoising procedure and they learned a generative model as the transition operator of Markov chain. 


\section{Face Synthesis based on Deep Learning}\label{deep}

\subsection{Convolutional Neural Networks}

Recent years have witnessed the significant breakthroughs of CNN in face recognition. Similarly, CNN has been widely applied to face synthesis. Taigman et al.\ \cite{taigman2014deepface} proposed a DeepFace system to obtain the human-level performance in face recognition and Yi et al.\ \cite{yi2014learning} presented a 11-layers convolutional neural network to learning face representation. A gradient ascent approach is deployed by Zhmoginov and Sandler \cite{zhmoginov2016inverting} to generate real-time face images. In order to recover high-quality facial pose, shape, expression, reflectance and illumination, Kim et al.\ \cite{kim2017inversefacenet} focused on an InverseFaceNet to obtain high-quality estimation. 

Without requiring extensive pose coverage in training data,  Peng et al.\ \cite{peng2017reconstruction} exerted a synthesis network to generate non-frontal view from a single frontal face.  Richardson et al.\ \cite{richardson2016learning} deployed an end-to-end convolutional neural network, which utilizes two parts to achieve transformation from coarse to fine. Capturing long-term dependencies along a sequence of transformations of shifts and rotations by recurrent structure, Yang et al.\ \cite{yang2015weakly} designed an end-to-end recurrent convolutional encoder-decoder network. Analogously inspired by encoder-decoder framework, Cole
et al.\ \cite{05} encoded the input face images into 1024-D feature vectors by FaceNet, and then used decoder network to generate the final images. Richardson et al.\ \cite{richardson20163d} put forward a CNN based approach which extracts the face geometry directly from original image. Zhang et al.\ \cite{zhang2016demeshnet} focused on inpainting corrupted face images through their DeMeshNet. They also proposed multi-task ConvNet \cite{zhang2016multi} with skip connection to further improve synthetic image quality of human face.



\subsection{Generative Adversarial Nets}
\label{gan}
Recently, Goodfellow et al.\ \cite{gan} proposed a novel framework for estimating generative models via an adversarial process called GAN for short. Soon much progress has been made in the fields of face synthesis by applying and altering GAN, and some state-of-art results have shown that their improved GAN models can produce synthesised frontal normal expression face images can even confuse normal human observers.

The theoretic basis mainly comes from decision theory and game theory.  GAN can be understood as a two-player non-cooperative game process. Generator and discriminator, the main components of GAN, are fighting against each other,  which can be formulated as the equation below: 
\begin{equation*}
\mathop {\min }\limits_G \mathop {\max }\limits_D V(D,G) = {\bm{E}_{x\sim{p_{data}}(x)}}[\log D(x)] + {\bm{E}_{x\sim{p_z}(z)}}[\log (1 - D(G(z)))]\\
\end{equation*}
where G is the generator and D is the discriminator, $p_{x}$ is the data distribution and $p_{z}$ is a known noise distribution. Satisfactory results in extensive experiments have shown that training the generator and discriminator alternatively through an iterative process with some stochastic gradient descent methods is experimentally effective .

Differing from traditional CNN, GAN is a generative model that can learn to fit the target data distribution. What's more, neither Markov chains nor inference is needed during learning, only back propagation is used to obtain gradients. However, original GAN suffers from some computational problems, e.g.\ the inferior performance caused by training generator too much without updating discriminator. Collapsed generator maps too many z to the same value of x and loses the capacity to fit the target data distribution. To address the aforementioned model collapse,  Zhao et al.\ \cite{ebgan} proposed energy based GAN (EBGAN) and viewed generator and discriminator as energy functions. Further,  Arjovsky et al.\ \cite{wgan} presented Wasserstein GAN (WGAN) based on Earth Mover distance. They proved that WGAN gets rid of the collapse problem to some extent.

\begin{figure}[htbp]
    \centering
    \begin{subfigure}[b]{\textwidth}
        \includegraphics[width=\textwidth]{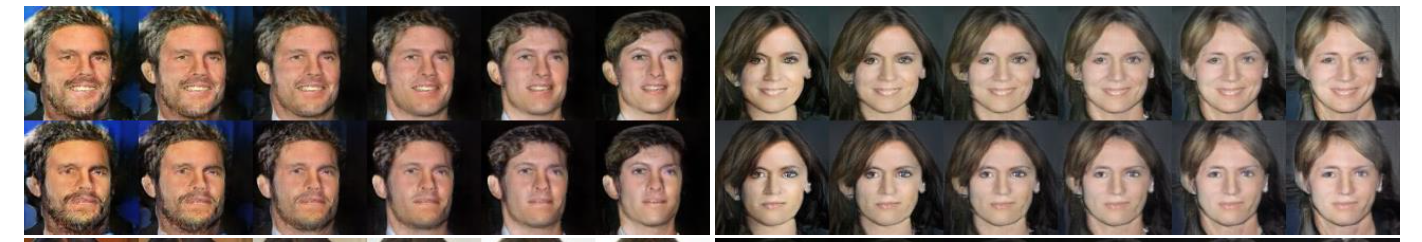}
        \caption{Results from \cite{012}, smiling attribute is added in the first row.}
        \label{couple}
    \end{subfigure}
    ~ 
    \begin{subfigure}[b]{\textwidth}
        \includegraphics[width=\textwidth]{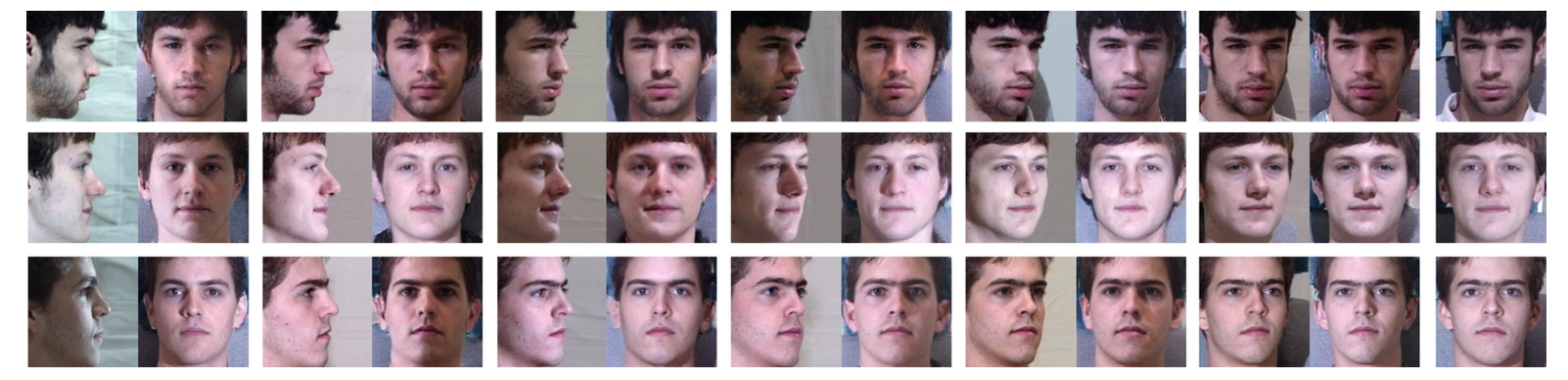}
        \caption{Face images that are in all kinds of pose synthesized by Huang's \cite{06} model.}
        \label{huang}
    \end{subfigure}
    ~
    \begin{subfigure}[b]{\textwidth}
        \includegraphics[width=\textwidth]{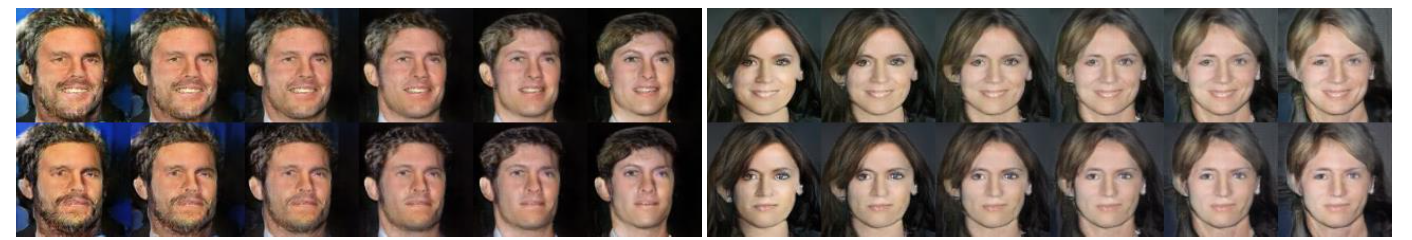}
        \caption{Two groups of images produced by acGAN \cite{010} that only differ in age.}
        \label{age}
    \end{subfigure}
    
    \caption{some synthesis results from \cite{010,06,012}}
    \label{syn}
\end{figure}

Current GAN models can handle the most headache cases, in which the PIE (pose, illumination and expression) of the input are under totally unconstrained situation, e.g.\ Radford et al.\ \cite{dcgan} designed a variation of GAN called DCGAN. Huang et al.\ \cite{06} focused on the local patches that has some semantic meaning and proposed TPGAN. Li et al.\ \cite{07} put their interest in the situation that parts of the face images are completely missing and came up with novel tow adversarial losses as well as a semantic parsing loss to complete the faces. \cite{08,lu2017conditional} applied an extension of GAN to a conditional setting. VariGANs model is proposed by Zhao et al.\ \cite{3} to solve the problem that generating multi-view images from only single view point. Tran et al.\ \cite{09} put forward DR-GAN which fuses the pose information and takes one or multiple face images with yaw angles as input to achieve pose invariant feature learning. Similarly, Antipov et al.\ \cite{010} concentrated on improving face synthesis in cross-age scenarios. Considering scene structure and context, Yang et al.\ \cite{18} presented LR-GAN that learns generated image background and foreground separately and recursively to produce a complete natural or face image. Figure \ref{syn} shows some synthesis results from \cite{010,06,012}.

Inspired by the capacity of fusing information of GAN, the application of face synthesis is also broadened. Kim et al.\ \cite{011} built a pair of dual models that are capable of two-way image domain transfer. Differing from Kim's model that can alter images semantically,  Liu and Tuzel \cite{012} proposed a model to produce images in two different domains simultaneously. Zhou et al.\ \cite{013} extracted feature representation with abstract semantic meanings and then ``cross-bred'' to finish domain transfer. Yin et al.\ \cite{16} presented a new method to learn to generate and modify the facial image coherently. A novel algorithm is put forward by \cite{17} to jointly learn latent codes for both identities and observations.

In general, methods based on GAN are basically on the behalf of state-of-the-art. Even human observers may be confused by synthetic face obtained from GAN model under some circumstances. However, ill-pose problem is yet an unresolved issue. It is worth mentioning that a framework combining 3DMM with GAN called FF-GAN was recently proposed by \cite{19}. In their work, 3DMM is utilized to provide shape and appearance priors to converge fleetly and they add a novel masked symmetry loss to recover visual quality under occlusions. 

\subsection{Other Deep Neural Networks}

Apart from CNN and GAN, lots of other deep neural network based methods have been developed. A novel face representation called the face identity-preserving (FIP) feature is proposed by \cite{zhu2013deep}. In order to better understand facial features and generate multi-view images, Zhu et al.\ \cite{zhu2014multi} proposed multi-view perceptron (MVP). In addition, their architecture has the ability to interpolate and synthesize the viewpoints that are not appeared in the training set. To avoid the high dependence on training set for AAM models, Duong et al.\ \cite{duong2016deep} proposed a Deep Appearance Models (DAMs) approach, which uses Deep Boltzmann Machine (DBM) to robustly capture the variations of facial shapes and appearances. 

So far, the popular methods of face synthesis have been reviewed from the viewpoint of model. To give a clear introduction of different applications and tasks in face synthesis, we tabulate some representative approaches in Table \ref{table1}.

\section{Public Available Databases and Evaluation Metrics}\label{database and metrics}

\subsection{Public Available Databases}
The means of acquiring data for subsequent face synthesis step vary with the specified methods. Some novel improvements on data-acquiring step may be considered as an commendable contribution for some researches. Moreover, to demonstrate the effectiveness of proposed methods, experimental results are analysed on the data collected from Internet or existing public databases. Those databases established originally for face detection, recognition or alignment are preferred because of provided high accuracy supervised information. Widely used databases are listed below:

Multi-PIE \cite{multi-pie}: The CMU Multi-PIE face database contains more than 750,000 images of 337 people. The database contains more than 305 GB of high resolution face images in total. The pose, illumination, and expression are the main interesting factors of the database. In total, 15 view points and 19 illumination and 7 expression conditions are recorded in controlled environment.

CASIA WebFace Database \cite{yi2014learning}: CASIA WebFace Database is a large scale dataset containing 10,575 subjects and 494,414 images. The collection process started from the well structured information in IMDB and then continued with the mean of web crawler. The faces in those images are detected and annotated automatically via clustering and face detection algorithm.

\afterpage{
\thispagestyle{empty}
\begin{table}[H]
\centering
\begin{tabular}{l p{210pt}}
\hline
\hline
Task Category & Representative Methods \\
\hline
\textbf{Different View Face Synthesis} &  \\
$\bullet{Frontal\ View\ Face\ Synthesis}$ & \emph{Generative Adversarial Nets}: Two-Pathway GAN (TP-GAN) \cite{06}, Conditional GAN (CGAN) \cite{08}, Face Frontalization GAN (FF-GAN) \cite{19}; \emph{Convolutional Neural Network}: InverseFaceNet \cite{kim2017inversefacenet,05}; \emph{Appearance Models}: Deep Appearance Models (DAMs) \cite{duong2016deep}; \emph{3D based Frontalization}: \cite{ferrari2016effective}; \emph{Statistic Learning based Method}: \cite{sagonas2015robust}; \emph{Sparse Representation}: \cite{zhao2014sparse};\\
$\bullet{Multipl\ Views\ Face\ Synthesis}$ & \emph{Generative Adversarial Nets}: Variational inference and the Generative Adversarial Networks (VariGANs) \cite{3}; \emph{Subspace Learning based Method}: Manifold based Method \cite{huang2010manifold, xu2013manifold}, Tensor-based Subspace Learning \cite{qiao2010tensor}; \emph{Statistic Learning based Method}: \cite{shim2010probabilistic}; \emph{Recurrent Convolutional Encoder-decoder Network}: \cite{yang2015weakly}; \emph{Other Deep Neural Net}: Multi-view Perceptron (MVP) \cite{zhu2014multi};\\
\textbf{Attributes Condition Face Synthesis} & \\
$\bullet{Semantic\ Attributes}$ & \emph{Generative Adversarial Nets}: Face Aging With Conditional GAN (Age-cGAN) \cite{010}, Semantically Decomposing GAN (SD-GAN) \cite{17}, Conditional CycleGAN \cite{lu2017conditional}, Disentangled Representation learning-GAN (DR-GAN) \cite{09}, Semi-latent GAN (SL-GAN) \cite{16}, GeneGAN \cite{013}; \emph{Subspace Learning based Method}: \cite{juefei2014facial};\\
$\bullet{Random\ Noise}$ & \emph{Generative Adversarial Nets}: GAN \cite{gan}, DCGAN \cite{dcgan}, Layered Recursive GAN (LRGAN) \cite{18}, Energy-based GAN (EBGAN) \cite{ebgan};\\
\textbf{Cross-Modality Face Synthesis} & \\
$\bullet{Face\ Sketch\ Synthesis}$ & \emph{Sparse Representation}: \cite{chang2010facenew,zhang2015facenew}; \emph{Structured Detail Enhancement}: \cite{song2016structured}; \emph{Random Sampling and Locality Constraint}: \cite{wang2017random}; \emph{Markov Random Fields (MRF) Model}: \cite{wangxiao2009face}; \emph{Embedded Hidden Markov Model (E-HMM)}: \cite{zhong2007facenew};\\
$\bullet{Cross-Domain\ Face\ Synthesis}$ & \emph{Generative Adversarial Nets}: DiscoGAN \cite{011}, CoGAN \cite{012};\\
\textbf{Super-Resolution Face Synthesis} & \emph{Appearance Models}: Deep Appearance Models (DAMs) \cite{duong2016deep}; \emph{Tensor based Method}: \cite{zhang2014face};\\
\textbf{Face Completion} & \emph{Generative Adversarial Nets}: Deep Generative Model \cite{07}; \emph{Convolutional Neural Network}: Multi-task ConvNet \cite{zhang2016multi}, DeMeshNet \cite{zhang2016demeshnet};\\
\textbf{3D Face Reconstruction} & \emph{Subspace Learning based Method}: Two-Fold Coupled Structure Learning (2FCSL) \cite{dou2014robust}, 3D Generic Elastic Model (3D-GEM) \cite{heo20123}, Gender and Ethnicity Specific GEMs (GE-GEMs) \cite{heo2012gender}; \emph{Statistical Learning based Method}: Cascaded Regression \cite{liu20153d, liu2016joint}; \emph{Geometry Modeling}: \cite{piotraschke2016automated, zhuang2008expressive}; \emph{Convolutional Neural Network}: \cite{richardson2016learning};\\
\hline
\hline
\end{tabular}
\caption{Taxonomy of Face Synthesis Approaches}
\label{table1}
\end{table}
}





CelebA \cite{liu2015faceattributes}: CelebFaces Attributes Dataset is a large-scale face attributes database that includes 10,177 identities, 202,599 face images and each image has 5 landmark locations and 40 binary attributes. In addition, this database covers large pose variations and background clutters.

VGG Face dataset \cite{vgg}: The dataset consists of 2.6M images of 2,622 identities. The images are collected from Internet using Google and Bing API via a series of filtering strategies including manual operations. The landmarks are obtained by manual annotation. The dataset has quite large images outside publicly available industrial datasets as well as substantial diversity in PIE (pose, illumination, expression).

LFW \cite{LFWTech, LFWTechUpdate}: Labeled Faces in the Wild \cite{LFWTech, LFWTechUpdate} is an unconstraint face recognition database. It collects 13,000 face images from the web and each face has a label. In addition, there are 1680 people have two or more distinct photos in database.

FaceWarehouse \cite{cao2014facewarehouse}: It is a 3D facial expression database for visual computing. It includes 150 people aged from 7 to 80 that have different ethnic backgrounds and each person has various expressions.

\begin{figure}[htbp]
    \centering
    \begin{subfigure}[b]{0.45\textwidth}
        \includegraphics[width=\textwidth,height=3in]{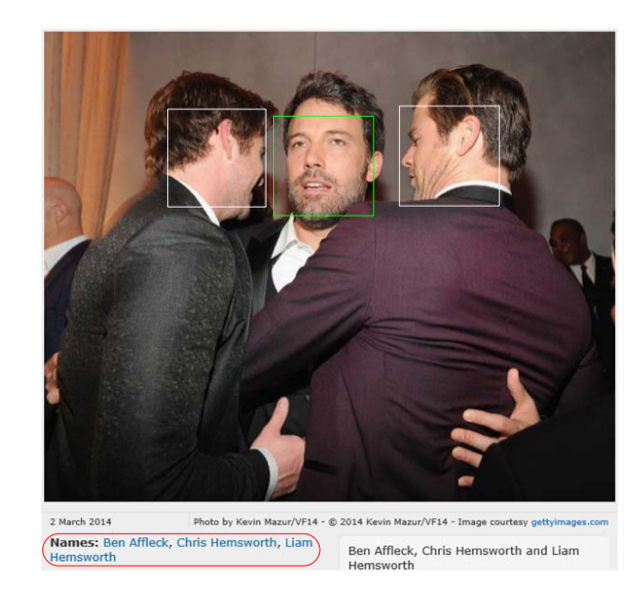}
        \caption{CASIA}
        \label{CASIA}
    \end{subfigure}
    ~ 
    \begin{subfigure}[b]{0.45\textwidth}
        \includegraphics[width=\textwidth,height=2.9in]{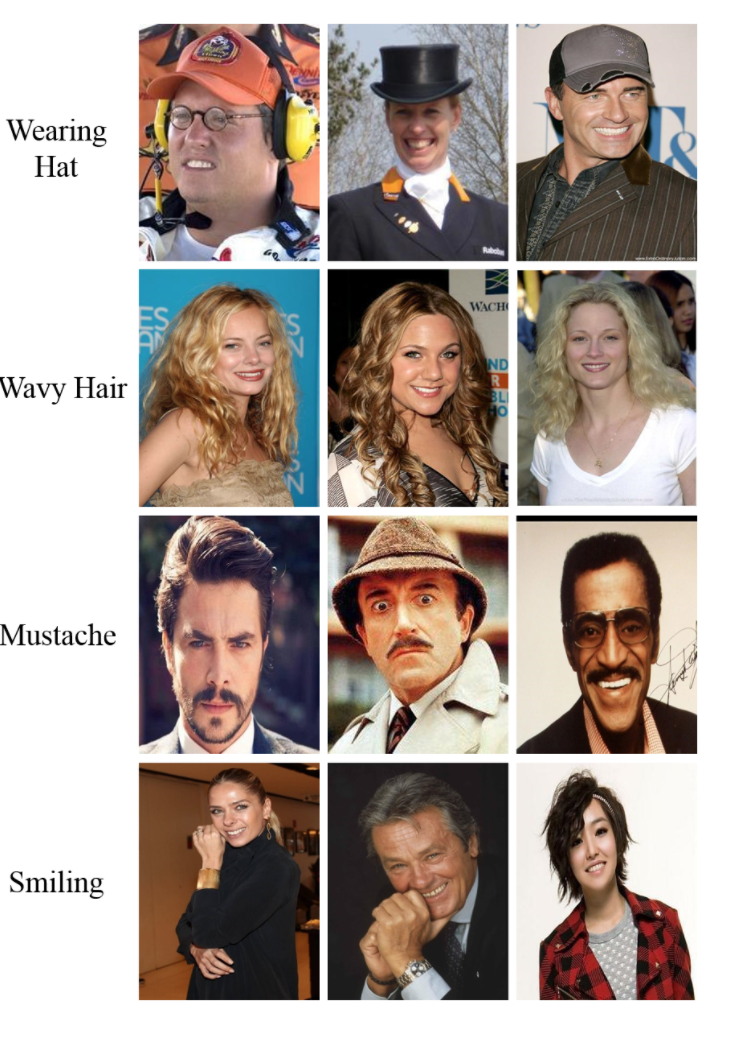}
        \caption{CelebA}
        \label{CelebA}
    \end{subfigure}
    \caption{some samples from the CASIA and CelebA databases}
    \label{database2}
\end{figure}

\begin{figure}[htbp]
    \centering
    \begin{subfigure}[b]{\textwidth}
        \includegraphics[width=\textwidth]{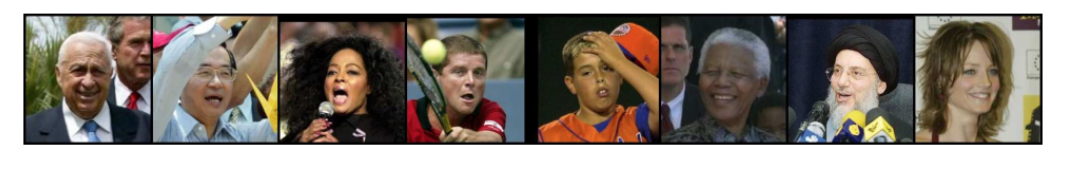}
        \caption{LFW}
        \label{LFW}
    \end{subfigure}
    ~ 
    \begin{subfigure}[b]{\textwidth}
        \includegraphics[width=\textwidth]{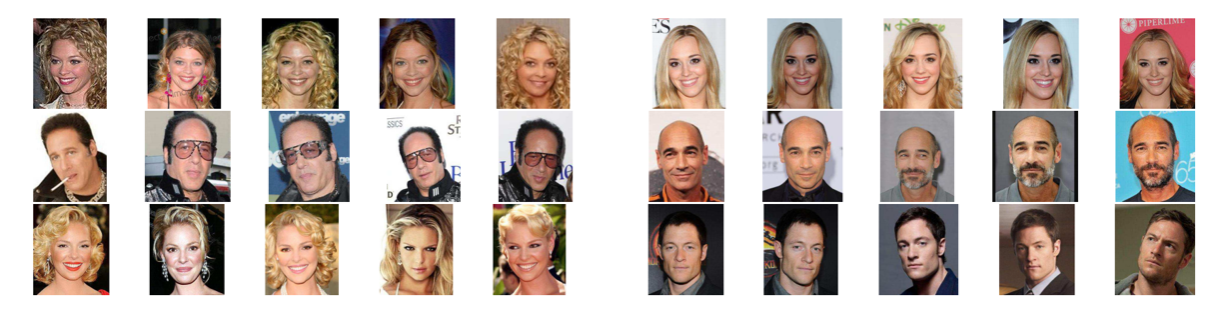}
        \caption{VGG}
        \label{vgg}
    \end{subfigure}
    ~
    \begin{subfigure}[b]{\textwidth}
        \includegraphics[width=\textwidth]{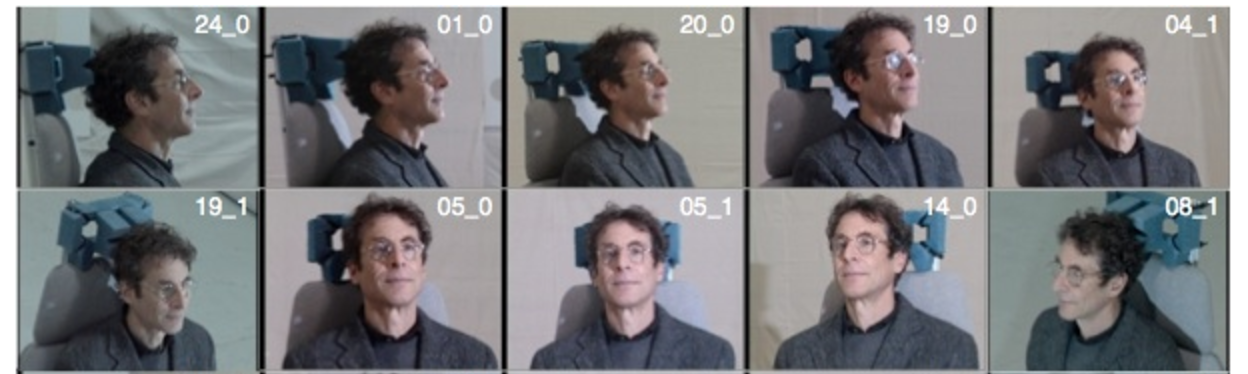}
        \caption{Multi-PIE}
        \label{multi-pie}
    \end{subfigure}
    ~
     \begin{subfigure}[b]{\textwidth}
        \includegraphics[width=\textwidth]{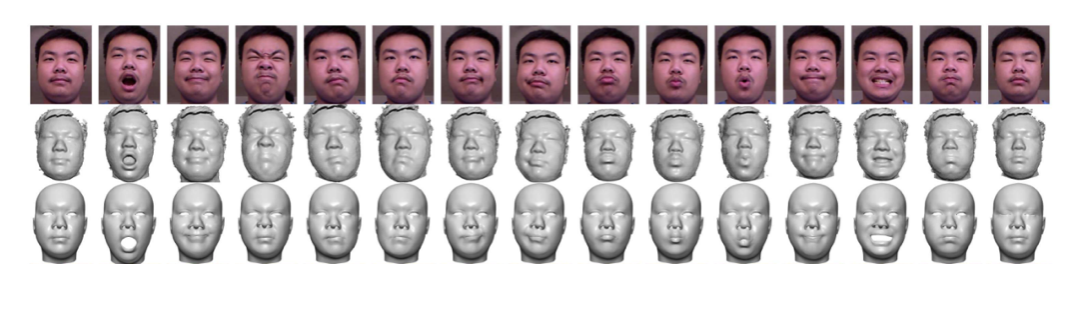}
        \caption{FaceWarehouse}
        \label{FaceWarehouse}
    \end{subfigure}
    
    \caption{more samples from the aforementioned databases}
    \label{database1}
\end{figure}

Some samples are shown in figures \ref{database2} and \ref{database1}.

\subsection{Evaluation Metrics}
The quality of face synthesis is a semantic concept which can not be accurately quantified. Employing human annotators to judge the visual quality of synthetic samples can only work for situation with very limited data amount. Till now, researchers are more likely to provide several best samples for reviewing. Pixel-level error is an optional choice, but judging the synthesized image quality merely by calculating the differences may produce results discrepant with the conclusion given by human annotators. Kim et al.\ \cite{kim2017inversefacenet} presented 3 error estimation methods, the first is photometric error and it computes the RMSE of RGB pixel values by comparing the input image and a rendering of the reconstructed face model. The second is geometric error that computes the RMSE of corresponding error in millimetre between the ground-truth geometry and their 3D model. The third is overlap of face masks and it estimates the intersection over union of face masks between their model and the input image. Besides, the improvement of synthetic samples for downstream algorithm dealing with some visual tasks face reconstruction can reflect the quality in some sense. As an alternative to human annotators, Salimans et al.\ \cite{salimans2017improved} proposed an automatic method to evaluate the performance of a GAN based model:
\begin{equation*}
Inception\ score=\exp ({\bm{E}_{\bm{x}}}KL(p(y|\bm{x})||p(y)))
\end{equation*}
The evaluation above is derived under the consideration that an admirable model should produce images that have close analogies to real data as many as possible, where x denotes a generated sample and y is the label predicted by a some out-of-the-box classifier. To measure the error in 3D model, Zhao et al.\ \cite{zhao2016mask} utilized the Euclidean distances to compute all the pairs of 3D points from ground truth model and their reconstructed model respectively.

\section{Conclusion}\label{Conclusion}

In this paper, we survey the recently representative methods in face synthesis. Existing face synthesis methods are roughly divided into two categories: traditional methods and deep learning models. GAN based on deep learning has become a popular generative model recently and simultaneously obtains rapid progress, so generating more realistic, diverse and high-quality face images becomes possible. Moreover, face synthesis based on 3D methods continues to be a hot topic because 3D methods model face directly from the view of geometry without large-scale data. Although great progress in face synthesis has been made, still there is a space for improvement and we propose the following promising directions as a guide for future development.
\begin{itemize}
\item \textbf{High-resolution face synthesis}: Notwithstanding GAN and other models have greatly improved the quality of face, high-resolution face synthesis is still an open problem. It is the most challenging yet significant for real-world practical deployments.
\item \textbf{Cross-modal face synthesis}: NIR imaging has been widely employed as a way to avoid illumination changes in outdoor circumstances. How to combine the advantages of VIS and NIR to improve face synthesis is an interesting topic. Incorporating 3D model to rectify face image is also a promising direction, but 3D reconsruction itself is still an intricate issue.
\item \textbf{Face systhesis via unpaired data}: It is expensive to collect large-scale paired data for face systhesis training. Hence how to use deep learning methods to learn from unpaired face data is an encouraging direction.
\item \textbf{Lack of databases}: Because of the limitation of current face databases, more large-scale multi-pose and in-the-wild face databases are urged to build for the improvement of face synthesis.
\item \textbf{Evaluation criterion}: Existing evaluation methods in face synthesis mainly contain two classes. One is evaluating indirectly by the improvement of down-stream algorithm that taking synthetic samples as input. Another method is to evaluate by human, which is a time-consuming and laborious work. How to design an objective, feasible, easily understanding and suitable calculating evaluation criterion remains challenging.
\end{itemize}

In the future, we would expect that more application of face synthesis will be developed and widely used in real-life. Furthermore, face synthesis will promote development of other tasks like face recognition, face detection and so on. Finally, the combination of several advanced techniques from multiple aspects is also helpful for face synthesis.

\section*{Acknowledgements}

This work is funded by the National Natural Science Foundation of China (Grant No. 61622310, 61473289), the Youth Innovation Promotion Association CAS (Grant No. 2015190), the Strategic Priority Research Program of the Chinese Academy of Sciences (Grant No. XDB02000000). 


\end{document}